\documentclass[sigconf]{acmart}
\usepackage{colortbl}
\usepackage{xcolor} 
\usepackage{makecell} 
\usepackage{amsmath}

\usepackage{amssymb}
\usepackage{booktabs}
\usepackage{enumitem}
\usepackage{multirow}
\usepackage{balance}
\AtBeginDocument{%
  }

\copyrightyear{2025}
\acmYear{2025}
\setcopyright{acmlicensed}
\acmConference[MM '25] {Proceedings of the 33rd ACM International Conference on Multimedia}{October 27--31, 2025}{Dublin, Ireland.}
\acmBooktitle{Proceedings of the 33rd ACM International Conference on Multimedia (MM '25), October 27--31, 2025, Dublin, Ireland}
\acmISBN{979-8-4007-2035-2/2025/10}
\acmDOI{10.1145/3746027.3755632}

\begin{document}

\title{CHORD: Customizing Hybrid-precision On-device Model for Sequential Recommendation with Device-cloud Collaboration}


\author{Tianqi Liu}
\orcid{0009-0002-7217-1290}
\authornote{Both authors contributed equally to this research.}
\affiliation{%
  \institution{Zhejiang University}
  \city{Hangzhou}
  \country{China}
}
\email{tqliu@zju.edu.cn}

\author{Kairui Fu}
\orcid{0009-0004-0284-671X}
\authornotemark[1]
\affiliation{%
  \institution{Zhejiang University}
  \city{Hangzhou}
  \country{China}
}
\email{fukairui.fkr@zju.edu.cn}

\author{Shengyu Zhang}
\orcid{0000-0002-0030-8289}
\authornote{Corresponding author.}
\affiliation{%
  \institution{Zhejiang University}
  \city{Hangzhou}
  \country{China}
}
\affiliation{%
  \institution{Shanghai Institute for Advanced Study of Zhejiang University}
  \city{Shanghai}
  \country{China}
}
\email{sy_zhang@zju.edu.cn}

\author{Wenyan Fan}
\orcid{0009-0006-5796-2229}
\affiliation{%
  \institution{Zhejiang University}
  \city{Hangzhou}
  \country{China}
}
\email{wenyan.17@outlook.com}

\author{Zhaocheng Du}
\orcid{0000-0002-1811-129X}
\affiliation{%
  \institution{Huawei Noah’s Ark Lab}
  \city{Shenzhen}
  \country{China}
}
\email{zhaochengdu@huawei.com}

\author{Jieming Zhu}
\orcid{0000-0002-5666-8320}
\affiliation{%
  \institution{Huawei Noah’s Ark Lab}
  \city{Shenzhen}
  \country{China}
}
\email{jamie.zhu@huawei.com}

\author{Fan Wu}
\orcid{0000-0003-0965-9058}
\affiliation{%
  \institution{Shanghai Jiao Tong University}
  \city{Shanghai}
  \country{China}
}
\email{fwu@cs.sjtu.edu.cn}

\author{Fei Wu}
\orcid{0000-0003-2139-8807}
\affiliation{%
  \institution{Zhejiang University}
  \city{Hangzhou}
  \country{China}
}
\email{wufei@zju.edu.cn}


\renewcommand{\shortauthors}{Tianqi Liu et al.}

\begin{abstract}
With the advancement of mobile device capabilities, deploying reranking models directly on devices has become feasible, enabling real-time contextual recommendations. When migrating models from cloud to devices, resource heterogeneity inevitably necessitates model compression. Recent quantization methods show promise for efficient deployment, yet they overlook device-specific user interests, resulting in compromised recommendation accuracy. While on-device finetuning captures personalized user preference, it imposes additional computational burden through local retraining.
To address these challenges, we propose a framework for \underline{\textbf{C}}ustomizing \underline{\textbf{H}}ybrid-precision \underline{\textbf{O}}n-device model for sequential \underline{\textbf{R}}ecommendation with \underline{\textbf{D}}evice-cloud collaboration (\textbf{CHORD}), leveraging channel-wise mixed-precision quantization to simultaneously achieve personalization and resource-adaptive deployment. CHORD distributes randomly initialized models across heterogeneous devices and identifies user-specific critical parameters through auxiliary hypernetwork modules on the cloud. Our parameter sensitivity analysis operates across multiple granularities (layer, filter, and element levels), enabling precise mapping from user profiles to quantization strategy. Through on-device mixed-precision quantization, CHORD delivers dynamic model adaptation and accelerated inference without backpropagation, eliminating costly retraining cycles. We minimize communication overhead by encoding quantization strategies using only 2 bits per channel instead of 32-bit weights. Experiments on three real-world datasets with two popular backbones (SASRec and Caser) demonstrate the accuracy, efficiency, and adaptivity of CHORD.
\end{abstract}




\maketitle

\section{Introduction}
\begin{figure*}[t]
    \centering
    \includegraphics[width=\textwidth]{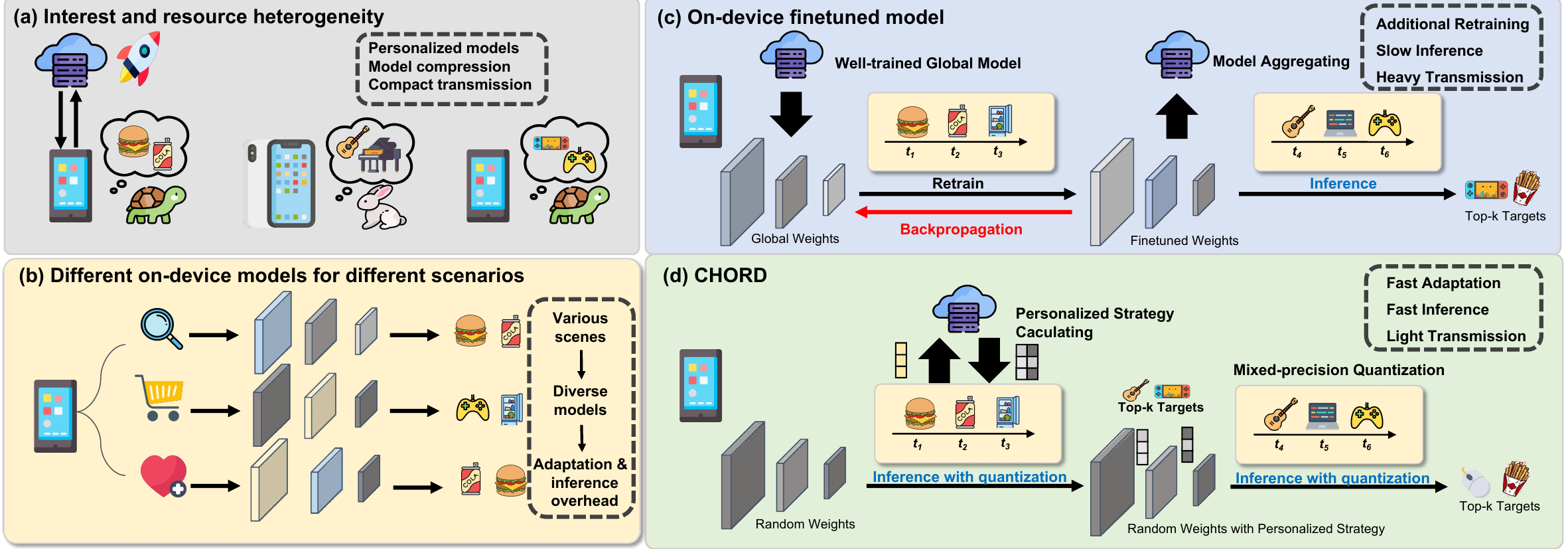}
    \caption{(a) Significant heterogeneity exists in interest patterns and computational resources between cloud and devices.
(b) Devices use multiple recommendation models to capture diverse interaction scenarios, further emphasizing the importance of fast adaptation and inference.
(c) Fine-tuned device models require costly retraining and backpropagation whenever user interests shift, forcing reliance on suboptimal models until these updates complete.
(d) In our "CHORD" approach, the cloud generates personalized channel-wise quantization strategies encoded as 2-bit representations upon interest shifts. Devices then utilize these quantized random-initialized models for efficient single-pass inference with improved accuracy.}
    \label{fig:wide_image1}
    \vspace{-2.5mm}
\end{figure*}

Nowadays, sequential recommendation has become a crucial component of recommendation systems, benefiting e-commerce, movies, music \cite{lei2021semi,zhao2019leveraging,hariri2012context,dai2024modeling,cai2021category,li2025mergenet,yu2025thinkrec} and many other domains. Representative models, including Caser \cite{tang2018personalized}, GRU4Rec \cite{jannach2017recurrent}, and SASRec \cite{kang2018self}, enhance the overall user experience through a sophisticated analysis of user behavior. Traditional recommendation systems predominantly rely on cloud-oriented data processing, which deploys a unified model on cloud servers for training and inference.
Despite its proven effectiveness, the required round-trip data transmission introduces response latency \cite{yin2024device,gong2020edgerec,fu2024diet}, making it difficult to capture real-time interests effectively. Furthermore, with billions of devices continuously interacting with cloud servers \cite{khan2019edge,lv2023duet}, the substantial bandwidth consumption presents a critical challenge for large-scale recommendation systems.

To alleviate these problems, on-device recommendation has emerged alongside the advancement of mobile devices in both hardware and software capabilities \cite{wang2020convergence}. 
Deploying reranking models directly on devices has become feasible, demonstrated by implementations on platforms Taobao \cite{gong2020edgerec} and Kuaishou \cite{gong2022real}. 
However, as depicted in Figure \ref{fig:wide_image1}(a), the resource heterogeneity \cite{khan2019edge,fu2025forward,wu2025knowledge} between devices and the cloud makes it impossible to directly deploy full-precision models on resource-constrained mobile devices, necessitating on-device model compression. 
Recently, quantization-based methods, especially mixed-precision quantization methods, have shown great promise in achieving efficient deployment \cite{bijl2024efficient,tang2024retraining,dong2019hawq,fu2023end,tang2022mixed}. By assigning different bit-widths to different layers or channels based on a uniform importance criterion, they strike a balance between model accuracy and inference efficiency.

Though effective, parameters seen as less important and quantized in lower bits could be merely incompatible with some devices, causing valuable information loss for others \cite{yao2021device}. Overlooking user interest heterogeneity, depicted in Figure \ref{fig:wide_image1}(a), leads to performance degradation \cite{qian2022intelligent,lv2024intelligent,yao2021device}. To achieve model customization, existing on-device fine-tuned approaches \cite{fast,yan2022device,nichol2018first} retrain their models locally, as is shown in Figure \ref{fig:wide_image1}(c), improving recommendation quality while incurring computational costs. Moreover, in sustainable deployment scenarios \cite{yin2024device}, the time-consuming model adaptation forces reliance on suboptimal models during model update sessions.

In light of these obstacles, our approach addresses two fundamental research challenges:

\vspace{-2mm}
\begin{enumerate}
    \item How to simultaneously achieve device-side customization and model compression, while enabling flexible adaptation across diverse device environments?
    \item How to minimize communication overhead, adaptation overhead, and inference overhead in device-cloud collaboration?
\end{enumerate}
\vspace{-2mm}

To address these challenges, we propose a lightweight and personalized recommendation framework called \textbf{CHORD}: \textbf{C}ustomizing \textbf{H}ybrid-precision \textbf{O}n-device model  for sequential \textbf{R}ecommendation with \textbf{D}evice-cloud  collaboration. We aim to simultaneously achieve model customization and resource-adaptive deployment, through channel-wise mixed precision quantization.

Inspired by the lottery ticket hypothesis \cite{ha2016hypernetworks}, CHORD distributes randomly initialized models across heterogeneous devices and identifies device-specific optimal quantization strategy. We view the process of discovering the ideal mixed-precision strategy, as finding the lottery ticket within the original model. To generate personalized strategy fitting to user interests, we leverage the rich computational resources of the cloud through multi-level user parameters saliency analysis. Meanwhile, on the device side, we apply personalized mixed-precision quantization to frozen layers, achieving efficient adaptation and inference. 

Consequently, we develop several sensitivity extractors on the cloud utilizing hypernetworks to generate multi-level parameter saliency metrics based on user profiles, while designing a user profiling generator on the device to capture real-time characteristics. Along with them, we implement a channel-wise strategy generator, considering layer, filter, and element level importance.
Filter-level importance establishes the foundation for our channel-wise quantization strategy, element-wise analysis provides weighted corrections to ensure richness of feature capturing, and inter-layer importance enables more comprehensive strategy formulation. In this process, we learn the mapping from heterogeneous user behaviors to compatible quantized structural representations.
Ultimately, devices will follow the encoded strategy and resource conditions to achieve personalized quantization (addressing challenge (1)).

Regarding communication overhead, we only need a 2-bit strategy encoding per output channel, compared to transmitting each weight element in 32-bit, dramatically reducing device-cloud communication costs. For adaptation overhead, we are capable of achieving personalized model adaptation by applying the quantization strategy with one forward pass. As for inference costs, devices can utilize the mixed-precision models to infer fast and accurately (addressing challenge (2)).

We conduct experiments on three real-world datasets and two widely used backbone networks, SASRec \cite{kang2018self} and Caser \cite{tang2018personalized} to demonstrate our accuracy, efficiency, and adaptivity. Our main contributions are as follows:
\begin{itemize}[leftmargin=*]
    \item We make an early attempt to propose a recommendation framework for device-cloud collaborative personalized mixed-precision quantization that generates lightweight compatible networks for heterogeneous devices with a forward pass.
    \item We generate personalized quantization strategies based on user interactions, achieving efficient transmission and flexible adaptation via compact strategy encoding and decoding mechanisms.
    \item We account for layer-wise, filter-wise, and element-wise parameter sensitivity when generating personalized strategies, resulting in improved recommendation performance.
    \item We validate our approach through extensive experiments and in-depth analysis on three real-world datasets, consistently outperforming other models. 
\end{itemize}

\section{RELATED WORK}
\subsection{On-device Recommendation}
On-device recommendation aims to provide real-time contextual recommendations. Some methods \cite{fast,yan2022device,nichol2018first} finetune the whole models with local samples, achieving model customization. However, the continuous evolution of user interests and resource bring great challenges to maintain  recommendation performance \cite{yin2024device}.
Researchers begin to leverage the computational resources on the cloud to alleviate these challenges.
The communication efficiency and model accuracy become their top priorities.
Some methods consider transmit partial or compressed weights.
DCCL \cite{yao2021device} incorporates meta-patch architecture to enable lightweight on-device personalization. 
ODUpdate \cite{xia2023towards} maintains efficiency by applying highly compressed parameter updates upon the existing model architecture. 
Other methods pay attention to the data distribution difference.
MPDA \cite{yan2022device} retrieves similar data from cloud to augment local device data, while some works prioritize model update scheduling \cite{qian2022intelligent,lv2024intelligent} by monitoring local data distribution shifts.
Our method harmonizes model compression with model personalization while achieving efficient device-cloud communication.
\subsection{Model Quantization}
Quantization navigates the tension between prediction accuracy and memory cost by representing each weight parameter with low-precision integers. 
Mixed-precision quantizations go one step further by assigning different levels of precision across layers or channels. These methods protect salient channels or layers, ensuring the preservation of critical information. Some methods use gradient-based optimization to find the optimal configurations. Bayesian \cite{van2020bayesian} decomposes of the quantization operation. DQ \cite{uhlich2019differentiable} learns the quantizer’s step size, dynamic range, and bitwidths upon gradient descent. Some approaches choose to use heuristic-based optimization. MPQ \cite{tang2022mixed} treats
the scale factors as importance indicators of a layer. HAWQ \cite{dong2019hawq} use the layer's Hessian spectrum as the importance metric. HAWQ-V2 \cite{dong2020hawq} further proves the effectiveness of the average Hessian trace. Other methods \cite{wang2019haq,ning2021simple,habi2020hmq} use metaheuristic or reinforcement learning to make a quantization strategy. Recently, adaptive quantization gains popularity because it can adapt to different resource conditions. AdaBits \cite{jin2020adabits} combines joint training with switchable clipping level technique
to enhance model quality. MBQuant \cite{zhong2025mbquant} utilizes a multi-branch topology to achieve adaptive deployment. Our method integrates mixed-precision quantization with adaptive quantization, enabling personalized quantization across heterogeneous devices.

\section{METHOD}
\begin{figure*}[t]
    \centering
    \includegraphics[width=\textwidth]{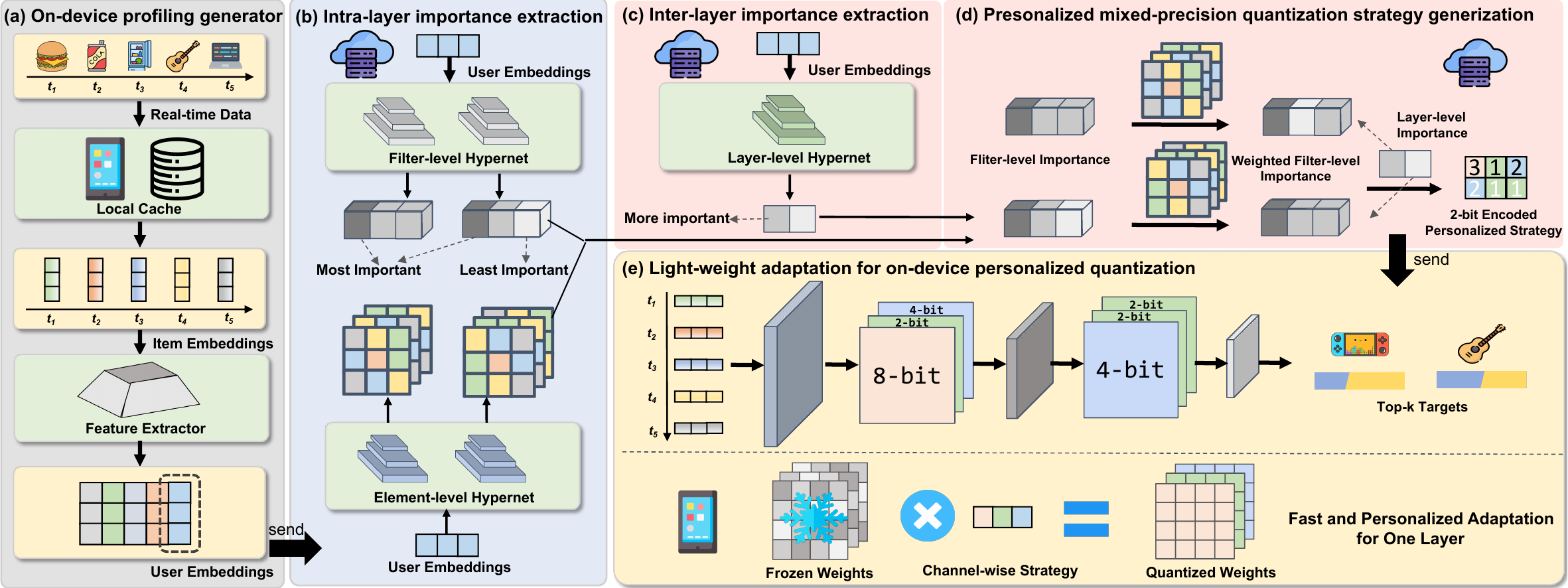}
    \caption{Overview of CHORD. (a) Devices will generate latent interest embeddings based on real-time interactions. (b) The cloud will discover filter-level and element-level relationships of parameters for each layer based on user profiles. (c) Another module on cloud will generate layer-level parameter sensitivity for each user. (d) The cloud will further utilize the element-level importance to reconstruct the filter-level importance. And then, the cloud will make a channel-wise quantization strategy based on the weighted filter-level importance and layer level importance. Transmission over the network only consists of 2-bit channel-wise strategy instead of weights. (e) Each device will share the same initial frozen weights. Devices will inference efficiently according to the customized mixed-precision quantization strategy with one forward pass. }
    \label{fig:wide_image2}
    \vspace{-2mm}
\end{figure*}
The general framework of our method is shown in Figure \ref{fig:wide_image2}. 

\subsection{Preliminary}

In our device-cloud collaborative recommendation  framework, we consider an environment with an item set $I = \{i_1, i_2, ..., i_n\}$ and a device set $D = \{d_1, d_2, ..., d_m\}$, where $n$ and $m$ represent the total number of items and devices, respectively. 
The cloud maintains access to historical interaction sequences $X^d_H = [x^d_1, x^d_2, ..., x^d_T]$ for each device $d \in D$, where each interaction $x^d_t \in I$ represents an item selected at a previous time step $t$. 
Separately, individual devices capture real-time interaction data $X^d_R = [x'^d_1, x'^d_2, ..., x'^d_k]$, representing the most recent user interactions. And devices hold a small number of candidate items embeddings (fewer than 100) \cite{gong2020edgerec} $I^d = [I^d_1, I^d_2, ..., I^d_p]$ sent by the cloud in each session for reranking tasks.
Our task is to predict the next clicked item $x'^d_{k+1}$ for each device. The limited bandwidth available for device-cloud communication is also a challenge to concern. Our overall research objective is to enhance on-device recommendation capabilities, enabling recommendations that are adaptive, personalized, and delivered with minimal latency.

Our primary objective can be formalized as:
\begin{equation}
\begin{aligned}
\max_{\sigma} & \sum_{d \in D} Q(\sigma_d, M)(X^d_R), \\
\text{subject to: } & B_d \leq B_{d\_max}, \quad C_d \leq C_{d\_max}.
\end{aligned}
\label{eq:eq1}
\end{equation}
The objective function in Equation \ref{eq:eq1} maximizes the utility of recommendations in all devices, where $\sigma_d$ is the personalized quantization strategy for device $d$. The function $Q$ applies personalized quantization to the model $M$ based on $\sigma_d$, producing a device-specific recommendation policy that processes real-time data $X^d_R$. The constraints enforce the bandwidth and computational limitations, where $B_d$ represents the bandwidth consumption of device $d$, $B_{d\_max}$ is its bandwidth limit, $C_d$ represents the real computational cost on device $d$, and $C_{d\_max}$ is its maximum allowable cost.

\subsection{Mixed-precision Quantization for Incompatible Parameters}
Network bandwidth and computational resources for devices continue to be constrained despite technological advances. The former one makes it costly for devices to frequently download comprehensive models from the cloud while the latter one necessities on-device model quantization. 
Fortunately, previous work \cite{frankle2018lottery,malach2020proving} has demonstrated the potential of random-initialized networks, showing that there exists an optimal subnetwork that can achieve comparable performance as training the entire network. Inspired by that, we propose to discover the effectiveness of finding the customized quantization strategy as finding the optimal subnets in recommender system.

In our device-cloud collaborative framework, each device adapts its model through applying the personalized quantization strategy with one forward pass. Consider that our backbone model SASRec~ \cite{kang2018self} consists of $K$ transformers, each with parameters $W_i = [w^0_i, w^1_i, ..., w^{p-1}_i]$, where $p$ is the number of linear layers in the $i$th transformer. For these transformers, the weights $W_i$ are frozen and the optimization process can be converted to finding an optimal quantization strategy for each linear layer. Similarly, we apply the customized quantization strategy to convolutional layers on backbone model Caser~ \cite{tang2018personalized} .
The quantization process for the entire model can be expressed as follows:
\begin{equation}
M_Q = Q(M, \sigma_d),
\label{eq:eq2}
\end{equation}
where $Q$ is the quantization function that applies the quantization strategy $\sigma_d$ to model $M$, and $\sigma_d$ represents the personalized quantization strategy for device $d$. 

For a weight tensor $W_i$ in a linear layer $i$ of model $M$, the quantization is performed channel-wise:
\begin{equation}
W_{i,k}^Q = Q_{\sigma_d(i,k)}(W_{i,k}),
\label{eq:eq3}
\end{equation}
where $W_{i,k}$ represents the $k$-th channel of layer $i$, and $Q_{\sigma_d(i,k)}$ is the quantization function with bit-width determined by $\sigma_d(i,k)$.

The personalized quantization strategy $\sigma_d$ is derived and encoded from the compact representation $\sigma_d'$ transmitted from the cloud:
\begin{equation}
\sigma_d = T(\sigma_d', R_d),
\label{eq:eq4}
\end{equation}
where $R_d$ represents the current resource conditions of device $d$ including computational capability,  battery level, etc. This adaptive transformation $T$ enables flexible adjustment of quantization intensity based on real-time device constraints.

This compact strategy representation $\sigma_d'$ uses only 2 bits per channel, dramatically reducing communication overhead compared to transmitting full-precision dense weights. 

The device-side fast adaptation mechanism further ensures optimal performance under varying resource conditions, making CHORD particularly suitable for large-scale recommendation systems with diverse devices.
\subsection{Device-specific Parameters Saliency Analysis}

To effectively generate device-specific quantization strategy, we need to precisely identify parameters critical to maintaining recommendation accuracy for individual users. 
Drawing inspiration from hypernetworks~ \cite{ha2016hypernetworks,von2019continual,navon2020learning,ruiz2024hyperdreambooth}, which were originally designed to generate weights for target networks, we leverage their inherent ability to learn complex mappings between different representation spaces. This architecture can be used to model the relationship between user latent representations and parameter sensitivity distributions. Through information transfer across model architecture, we can effectively solve our parameter sensitivity analysis task.
\begin{equation}
\alpha = H(z), \label{eq:alpha_generation}
\end{equation}
\noindent where $H$ is the hypernet to identify salient parameters, $z$ represents user latent interest embeddings, and $\alpha$ denotes the 
personalized parameter sensitivity.

\subsubsection{User profiling generation}
To generate user latent interest embeddings, we need to analyze user real-time interactions. To keep the information up to date and identify the scheduling of updating models, we adopt a lightweight sequence extractor GRU $\mathcal{G}_d$  on device for generating compact user representations: 
\begin{equation}
z_d = \mathcal{G}_d(X^d_H).
\end{equation}
This extractor transforms item sequences into a user embedding vector $z_d$, yielding a $l$-dimensional representation rather than an $k \times l$ matrix, which facilitates efficient processing of user preference.

\subsubsection{Multi-granularity sensitivity extraction}
Parameter sensitivity varies at different structural levels within a neural network. We analyze this sensitivity on cloud at three distinct levels: layer-level, filter-level, and element-level. 
This hierarchical  approach ensures that computational resources are allocated where they provide the greatest impact on model performance.

1. \textbf{Filter-level Sensitivity}: 
For each frozen layer, we use a filter-level hypernet to capture the importance of filter:
\begin{equation}
\alpha^F_i = H^F_i(z),
\end{equation}
where $\alpha^F_i \in \mathbb{R}^{d_{out}}$ represents the importance of each output channel in layer $i$, and $H^F_i$ is the filter-level hypernetwork. This forms the foundation for our channel-wise quantization approach.

2. \textbf{Element-level Sensitivity}: For each frozen layer, we use an element-level hypernet to capture the importance of parameter:
\begin{equation}
\alpha^E_i = H^E_i(z),
\end{equation}
where $\alpha^E_i$ assigns importance scores to each weight element in layer $i$, providing weighted refinements to enhance the precision of filter-level representations.

3. \textbf{Layer-level Sensitivity}: We use a element-level hypernet to capture importance across all layers:
\begin{equation}
\alpha^L = H^L(z),
\end{equation}
where $\alpha^L$ determines the global importance distribution across the network architecture, enabling the identification of sensitive layers and ensuring balanced performance across the entire model.

\subsection{Personalized Strategy Generation for Mixed-precision Quantization }
\subsubsection{Channel-wise strategy generation}
Channel-wise quantization strategies that rely solely on filter-level importance cannot capture the internal distribution of weight values, leading to suboptimal bit allocation. Inspired by recent works \cite{fu2024diet} that utilize both filter-level and element-level importance to capture more comprehensive information, we construct the weighted filter-level importance to better understand the parameter sensitivity.
To map element-level importance to channel-level metrics, we aggregate the element-wise importance scores using L1 distance:
\begin{equation} 
S_{i,j} = \sum_{k \in C_j} |\alpha^E_{i,k}|,
\end{equation}
where $C_j$ denotes the set of elements in channel $j$, and $S_{i,j}$ represents the aggregated importance. 

For original filter-level importance, we apply softmax normalization to ensure proper weighting. The weighted channel sensitivity then combines both granularities through multiplication:
\begin{equation}
\alpha^W_{i,j} = \text{softmax}(\alpha^F_i) \cdot S_{i,j}.
\end{equation}

This integration ensures we prioritize channels with both high contextual relevance (filter-level) and significant internal weight distribution (element-level). 
Based on the weighted channel sensitivity $\alpha^W_{i,j}$ , we define a quantization strategy transformation function $\Gamma$ that maps sensitivity values to discrete bit-width allocations and encode them into 2-bits per channel:
\begin{equation}
\sigma_d' = \Gamma(\alpha^W_{i,j}, \beta),
\end{equation}
where $\sigma_d'$ represents the final encoded quantization strategy transmitted to the device. Through $\Gamma$, channels are categorized into three tiers using a single hyperparameter $\beta$: in our settings, those with highest sensitivity (top $\beta\%$) receive 8-bit precision, channels with moderate sensitivity (next $\beta\%$) receive 4-bit precision, and the remaining channels are allocated 2-bit precision.

To address the non-differentiable nature of this mapping operation during training, we employ a straight-through estimator that maintains discrete bit allocation in the forward pass while allowing gradient flow in the backward pass.

\subsubsection{layer-wise strategy improvement}
Existing mixed-precision quantization approaches mainly focus on assigning different bit-widths to different layers \cite{tang2022mixed,dong2019hawq}. They acknowledge that layers contribute differently to overall model performance, inspiring us not to treat intra-level importance in isolation. Thus, we utilize our layer-level importance scores $\alpha^L$ to identify particularly sensitive and insensitive layers. Through function $\Lambda$, we extract these critical layers and apply special bit-width adjustments:
\begin{equation}
\sigma_d' = \Lambda(\alpha^L, \sigma_d'),
\end{equation}
where $\sigma_d'$ represents our refining quantization strategy. The most sensitive layers receive an additional precision boost, elevating their quantization to more precise representations. Conversely, the least sensitive layers are further compressed. 

The refined strategy $\sigma_d'$ now incorporates both inter-layer and intra-layer sensitivity, generating a comprehensive mixed-precision schema across the four quantization levels (e.g. 2, 4, 6, and 8 bits) which successfully optimizes the precision-efficiency trade-off for personalized model deployment.

\section{EXPERIMENTS}

\subsection{Experimental Setup}

\subsubsection{Datasets}
\begin{table}
\caption{Statistics of datasets.}
\vspace{-3mm}
\label{tab:datasets}
\begin{tabular}{p{1.5cm}cccc}
\toprule
\textbf{Dataset} & \textbf{\#Users} & \textbf{\#Items} & \textbf{\#Interactions} & \textbf{\#Density} \\
\midrule
CD & 31,482 & 68,307 & 867,853 & 0.040\% \\
Yelp & 97,052 & 94,279 & 2,943,170 & 0.032\% \\
ML-100K & 943 & 928 & 94,672 & 10.802\% \\
\bottomrule
\end{tabular}
\vspace{-5mm}
\end{table}

The experiments are conducted on three datasets: Amazon-CD\footnote{https://nijianmo.github.io/amazon/index.html}, Yelp\footnote{https://www.yelp.com/dataset/challenge} and MovieLens-100k\footnote{https://grouplens.org/datasets/movielens/100k}. Detailed statistics of them are shown in Table ~\ref{tab:datasets}. To maintain data quality, we implement a 10-core setting where both users and items with fewer than 10 interactions are excluded from all datasets. For each user, we chronologically order their interactions and allocate the most recent one to the test set, while others serve as training data.

\subsubsection{Baselines}
We adopt SASRec~ \cite{kang2018self}, Caser~ \cite{tang2018personalized} as the base models consisting of two popular architectures transformer and CNN. We also include seven baselines applied on them: (1) \textbf{Traditional Methods}: On-device Static and On-device Finetune (Static and Finetune). (2) \textbf{Compression-based Methods}: PEMN \cite{bai2022parameter}, Quant \cite{krishnamoorthi2018quantizing}, AdaBits~ \cite{jin2020adabits}, 
RFQuant~ \cite{tang2024retraining},
MBQuant~ \cite{zhong2025mbquant}. 
\begin{itemize}[leftmargin=*]
    \item \textbf{PEMN}~ \cite{bai2022parameter} They early attempt to validate the potential of fixed weights with limited unique values by learning weight masks.
    \item \textbf{Quant}~ \cite{krishnamoorthi2018quantizing} They explore the effectiveness of a uniform affine quantization scheme utilizing per-channel quantization for weights and per-layer quantization for activations.
    \item \textbf{AdaBits}~ \cite{jin2020adabits} They explore using adaptive bit-widths for adaptively deploying on-device models, balancing accuracy against efficiency.
    \item \textbf{RFQuant}~ \cite{tang2024retraining} They develop a bit-width scheduler that progressively freezes the most unstable bit-widths during the training process, ensuring proper convergence for the remaining bit-width parameters.
    \item \textbf{MBQuant}~ \cite{zhong2025mbquant} They employs a multi-branch topology that uses fixed 2-bit weight quantization across independent branches, reducing quantization errors through strategic branch selection.
\end{itemize}

\subsubsection{Evaluation Metrics}
We primarily focus on model accuracy, model inference efficiency and device-cloud communication overhead. 
To assess recommendation quality, we employ two commonly adopted metrics: NDCG and Hit rate. Higher values indicate superior recommendation performance. We use average bits to measure inference efficiency. Lower values represent faster inference. Regarding communication efficiency, we use the million bit count of model parameters (Param) that are transmitted. A lower Param value correlates with lighter transmission overhead.

\subsection{Overall Performance}
\begin{table*}[t]
\caption{Overall Performance on recommendation accuracy and resource overhead. }
\label{tab:performance}
\centering

\definecolor{lightblue}{rgb}{0.85,0.95,1.0} 
\resizebox{\textwidth}{!}{
\begin{tabular}{c|c|c|ccccc|ccccc|ccccc}
\toprule[1.5pt]
\multirow{2}{*}{\textbf{Model}} & \multirow{2}{*}{\textbf{Method}} & \multirow{2}{*}{\textbf{Avg Bits}} & \multicolumn{5}{c|}{\textbf{CD}} & \multicolumn{5}{c|}{\textbf{Yelp}} & \multicolumn{5}{c}{\textbf{ML-100K}} \\
\cline{4-18}
 & & & \textbf{NDCG@5} & \textbf{HR@5} & \textbf{NDCG@10} & \textbf{HR@10} & \textbf{Param} & \textbf{NDCG@5} & \textbf{HR@5} & \textbf{NDCG@10} & \textbf{HR@10} & \textbf{Param} & \textbf{NDCG@5} & \textbf{HR@5} & \textbf{NDCG@10} & \textbf{HR@10} & \textbf{Param} \\
\midrule
\multirow{8}{*}{\textbf{Caser}} & Base & 32 & 0.0183 & 0.0253 & 0.0216 & 0.0356 & 0.4968 & 0.0097 & 0.0157 & 0.0132 & 0.0266 & 0.4968 & 0.0439 & 0.0668 & 0.0599 & 0.1177 & 0.4968 \\
 & Finetune & 32 & 0.0184 & 0.0254 & 0.0217 & 0.0358 & 0.4968 & 0.0097 & 0.0157 & 0.0132 & 0.0265 & 0.4968 & 0.0448 & 0.0679 & 0.0607 & 0.1188 & 0.4968 \\
 & PEMN & 3$\dagger$ & 0.0095 & 0.0134 & 0.0120 & 0.0211 & 0.0336 & 0.0068 & 0.0109 & 0.0094 & 0.0189 & 0.0336 & 0.0283 & 0.0488 & 0.0421 & 0.0923 & 0.0336 \\
 & Quant & 3 & 0.0090 & 0.0133 & 0.0109 & 0.0191 & 0.0490 & 0.0066 & 0.0107 & 0.0092 & 0.0188 & 0.0490 & 0.0217 & 0.0371 & 0.0298 & 0.0615 & 0.0490 \\
 & AdaBits & 3 & 0.0067 & 0.0103 & 0.0084 & 0.0157 & 0.0490 & 0.0062 & 0.0100 & 0.0087 & 0.0180 & 0.0490 & 0.0335 & 0.0520 & 0.0468 & 0.0944 & 0.0490 \\
 & RFQuant & 3 & 0.0061 & 0.0088 & 0.0077 & 0.0136 & 0.0490 & 0.0035 & 0.0056 & 0.0050 & 0.0101 & 0.0490 & 0.0278 & 0.0467 & 0.0374 & 0.0764 & 0.0490 \\
 & MBQuant & 3 & 0.0149 & 0.0211 & 0.0181 & 0.0310 & 0.0490 & 0.0075 & 0.0124 & 0.0105 & 0.0216 & 0.0490 & 0.0347 & 0.0594 & 0.0461 & 0.0954 & 0.0490 \\
 & CHORD & 3 & \cellcolor{lightgray}\textbf{0.0298} & \cellcolor{lightgray}\textbf{0.0362} & \cellcolor{lightgray}\textbf{0.0332} & \cellcolor{lightgray}\textbf{0.0468} & \cellcolor{lightgray}\textbf{0.0029} & \cellcolor{lightgray}\textbf{0.0104} & \cellcolor{lightgray}\textbf{0.0168} & \cellcolor{lightgray}\textbf{0.0143} & \cellcolor{lightgray}\textbf{0.0290} & \cellcolor{lightgray}\textbf{0.0029} & \cellcolor{lightgray}\textbf{0.0493} & \cellcolor{lightgray}\textbf{0.0753} & \cellcolor{lightgray}\textbf{0.0639} & \cellcolor{lightgray}\textbf{0.1220} & \cellcolor{lightgray}\textbf{0.0029} \\
\cline{2-18}
 & \multicolumn{2}{c|}{Improvement} & \cellcolor{lightblue}62.8\% & \cellcolor{lightblue}43.1\% & \cellcolor{lightblue}53.7\% & \cellcolor{lightblue}31.5\% & \cellcolor{lightblue}$\times$ 173.8 & \cellcolor{lightblue}7.2\% & \cellcolor{lightblue}7.0\% & \cellcolor{lightblue}8.3\% & \cellcolor{lightblue}9.0\% & \cellcolor{lightblue}$\times$ 173.8 & \cellcolor{lightblue}12.3\% & \cellcolor{lightblue}12.7\% & \cellcolor{lightblue}6.7\% & \cellcolor{lightblue}3.7\% & \cellcolor{lightblue}$\times$ 173.8 \\
\midrule
\multirow{8}{*}{\textbf{SASRec}} & Base & 32 & 0.0258 & 0.0320 & 0.0293 & 0.0431 & 3.9936 & 0.0107 & 0.0172 & 0.0145 & 0.0292 & 2.6624 & 0.0342 & 0.0551 & 0.0515 & 0.1092 & 2.6624 \\
 & Finetune & 32 & 0.0257 & 0.0320 & 0.0294 & 0.0433 & 3.9936 & 0.0106 & 0.0171 & 0.0146 & 0.0294 & 2.6624 & 0.0344 & 0.0541 & 0.0526 & 0.1103 & 2.6624 \\
 & PEMN & 3$\dagger$ & 0.0375 & 0.0454 & 0.0412 & 0.0566 & 0.3072 & 0.0115 & 0.0186 & 0.0155 & 0.0312 & 0.2048 & 0.0429 & 0.0785 & 0.0601 & 0.1326 & 0.2048 \\
 & Quant & 3 & 0.0157 & 0.0210 & 0.0184 & 0.0294 & 0.4301 & 0.0084 & 0.0134 & 0.0115 & 0.0233 & 0.2867 & 0.0317 & 0.0530 & 0.0479 & 0.1029 & 0.2867 \\
 & AdaBits & 3 & 0.0017 & 0.0029 & 0.0021 & 0.0042 & 0.4301 & 0.0061 & 0.0100 & 0.0083 & 0.0169 & 0.2867 & 0.0336 & 0.0541 & 0.0473 & 0.0976 & 0.2867 \\
 & RFQuant & 3 & 0.0033 & 0.0046 & 0.0040 & 0.0067 & 0.4301 & 0.0040 & 0.0065 & 0.0054 & 0.0109 & 0.2867 & 0.0276 & 0.0477 & 0.0463 & 0.1071 & 0.2867 \\
 & MBQuant & 3 & 0.0293 & 0.0361 & 0.0326 & 0.0465 & 0.4301 & 0.0105 & 0.0170 & 0.0143 & 0.0289 & 0.2867 & 0.0353 & 0.0562 & 0.0500 & 0.1018 & 0.2867 \\
 & CHORD & 3 & \cellcolor{lightgray}\textbf{0.0370} & \cellcolor{lightgray}\textbf{0.0453} & \cellcolor{lightgray}\textbf{0.0412} & \cellcolor{lightgray}\textbf{0.0584} & \cellcolor{lightgray}\textbf{0.0653} & \cellcolor{lightgray}\textbf{0.0118} & \cellcolor{lightgray}\textbf{0.0188} & \cellcolor{lightgray}\textbf{0.0160} & \cellcolor{lightgray}\textbf{0.0318} & \cellcolor{lightgray}\textbf{0.0435} & \cellcolor{lightgray}\textbf{0.0519} & \cellcolor{lightgray}\textbf{0.0870} & \cellcolor{lightgray}\textbf{0.0707} & \cellcolor{lightgray}\textbf{0.1463} & \cellcolor{lightgray}\textbf{0.0435} \\
\cline{2-18}
 & \multicolumn{2}{c|}{Improvement} & \cellcolor{lightblue}43.4\% & \cellcolor{lightblue}41.6\% & \cellcolor{lightblue}40.6\% & \cellcolor{lightblue}35.5\% & \cellcolor{lightblue}$\times$ 61.20 & \cellcolor{lightblue}10.3\% & \cellcolor{lightblue}9.3\% & \cellcolor{lightblue}10.3\% & \cellcolor{lightblue}8.9\% & \cellcolor{lightblue}$\times$ 61.20 & \cellcolor{lightblue}51.8\% & \cellcolor{lightblue}57.9\% & \cellcolor{lightblue}37.3\% & \cellcolor{lightblue}34.0\% & \cellcolor{lightblue}$\times$ 61.20 \\
\bottomrule[1.5pt]
\end{tabular}
}
\begin{flushleft}
\footnotesize{$^\dagger$ denotes methods that achieve equivalent weight sparsity through non-quantization techniques.}
\end{flushleft}
\end{table*}
Table~\ref{tab:performance} presents a comparative analysis of CHORD and seven baseline methods. Our experimental results demonstrate that CHORD consistently outperforms all compared methods.

When examining the traditional methods, we observe that on-device Finetune slightly improves over the Base model at the cost of increased local computation overhead. This marginal enhancement suggests that fine-tuning alone is insufficient for on-device recommendation scenarios where computational resources are constrained and user samples are limited.

When compared to the compression-based methods, we find that the reduction of valuable parameters deeply affect the recommendation accuracy. PEMN, which applies the lottery ticket hypothesis for network pruning, shows mixed results. While it achieves notable improvements with SASRec, it underperforms on Caser across all datasets. This inconsistency indicates that PEMN struggles to identify effective personalized subnetworks across different architectures, limiting its generalizability. 

Regarding quantization approaches, standard Quant methods show considerable performance degradation. For Caser on CD, Quant reduces NDCG@5 from 0.0183 to 0.0090, demonstrating the difficulty in preserving recommendation accuracy while reducing inference efficiency. RFQuant, despite its bit-width scheduling approach, shows even worse performance than Quant on Caser, indicating that progressive bit freezing and choosing may not be well-suited for recommendation models. AdaBits, which attempts to balance accuracy and efficiency through adaptive bit-widths, does not guarantee a improvement across datasets, suggesting that its adaptive strategy can potentially 
require more time and resources to learn. MBQuant demonstrates better results than Quant by employing multi-branch topology, yet still falls short of matching the Finetune performance due to the ignorance of user-specific features.

CHORD consistently outperforms all baselines across datasets and model architectures while achieving remarkable inference and communication efficiency. Let alone the inference speed up with 3-bit mixed quantization, for Caser, CHORD improves NDCG@5 by up to 62.8\% on CD while reducing transmission parameters by 173.8$\times$. Similarly, with SASRec, CHORD enhances performance by up to 51.8\% on ML-100K with a 61.2$\times$ parameter reduction. These results demonstrate that CHORD effectively balances recommendation accuracy, inference efficiency and transmission overhead, making it particularly suitable for resource-constrained on-device deployment.

\subsection{Ablation Study}
\begin{table}[h]
  \caption{Ablation Study}
  \vspace{-2mm}
  \label{tab:ablation}
  \resizebox{\linewidth}{!}{
      \begin{tabular}{c|c|cc|cc}
    \toprule[1.5pt]
    \multirow{2}{*}{\textbf{Model}} & \multirow{2}{*}{\textbf{Method}} & \multicolumn{2}{c|}{\textbf{ML-100K}} & \multicolumn{2}{c}{\textbf{Yelp}}  \\
    \cmidrule(lr){3-4}\cmidrule(lr){5-6}
    & & \textbf{NDCG@10} & \textbf{HR@10} & \textbf{NDCG@10} & \textbf{HR@10} \\
    \midrule
    \multirow{4}{*}{\textbf{Caser}} & Quant & 0.0298 & 0.0615 & 0.0092 & 0.0188 \\
    & +Customization & 0.0569 & 0.1156 & 0.0141 & 0.0284 \\
    & +Weighted Channel & 0.0587 & 0.1198 & 0.0141 & 0.0287 \\
    & CHORD & 0.0639 & 0.1220 & 0.0143 & 0.0290 \\
    \midrule
    \multirow{4}{*}{\textbf{SASRec}} & Quant & 0.0479 & 0.1029 & 0.0115 & 0.0233 \\
    & +Customization & 0.0687 & 0.1432 & 0.0157 & 0.0314 \\
    & +Weighted Channel & 0.0699 & 0.1421 & 0.0159 & 0.0319 \\
    & CHORD & 0.0707 & 0.1463 & 0.0160 & 0.0318 \\
    \bottomrule[1.5pt]
  \end{tabular}

  }

\end{table}
To understand the contribution of each component in our proposed CHORD framework, we conduct a comprehensive ablation study. We incrementally add components to a baseline quantization model and evaluate performance on two datasets using two recommendation backbone models. Table~\ref{tab:ablation} presents the results of our experiments.

\begin{itemize}[leftmargin=*]
    \item \textbf{Quant} employs standard min-max quantization uniformly across the model without personalization. This serves as our baseline and represents the conventional approach to model compression in resource-constrained environments.
    
    \item \textbf{+Customization} introduces user-specific quantization strategies through filter-level hypernetworks. This component enables personalized bit allocation based on user interaction patterns, but treats each channel as an independent unit without considering internal weight distributions.
    
    \item \textbf{+Weighted Channel} enhances the filter-level importance scores by incorporating element-level sensitivity information. By aggregating fine-grained weight importance within each channel, this component captures the interdependencies between weights and produces more informed quantization decisions. The performance gains are particularly evident in the transformer-based SASRec architecture.
    
    \item \textbf{CHORD} represents our complete framework with the addition of salient layer improvement. This final component identifies particularly sensitive and insensitive layers, enabling adaptive precision allocation across the model architecture. The  quantization strategy achieves consistent improvements, especially on ML-100K dataset, benefiting from comprehensive sensitivity analysis.
\end{itemize}

\subsection{In-depth Analysis}
\subsubsection{Detailed analysis on training performance}
\begin{figure}
    \centering
    \includegraphics[width=1\linewidth]{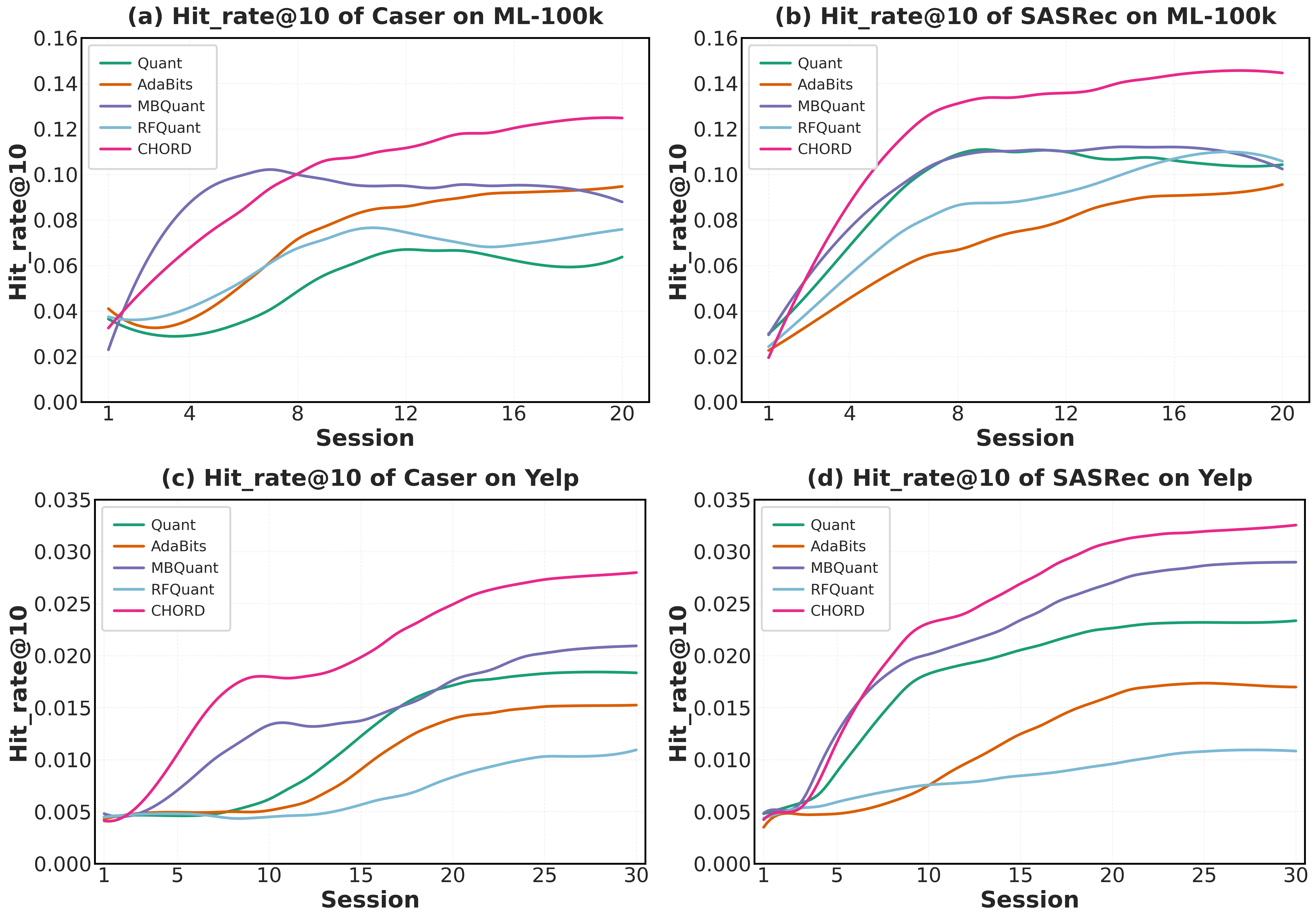}
    \caption{Detailed Training Analysis Compared to four quantization baselines on ML-100K and Yelp}
    \label{fig:training}
    \vspace{-4mm}
\end{figure}
To further investigate the effectiveness of CHORD, we plot the Hit\_rate@10 progression during training compared to four quantization-based methods in Figure~\ref{fig:training}. The results demonstrate that our personalized quantization approach consistently outperforms baseline methods across different datasets and model architectures. 
We observe that quantization methods like Quant and RFQuant often struggle with performance oscillations, particularly evident in the Caser model. While MBQuant shows competitive performance in certain settings, its fixed quantization strategy lacks the adaptability provided by CHORD's multi-granularity importance extraction. 

\subsubsection{Sensitivity analysis on channel selection threshold}
\begin{figure}
    \centering
    \includegraphics[width=1\linewidth]{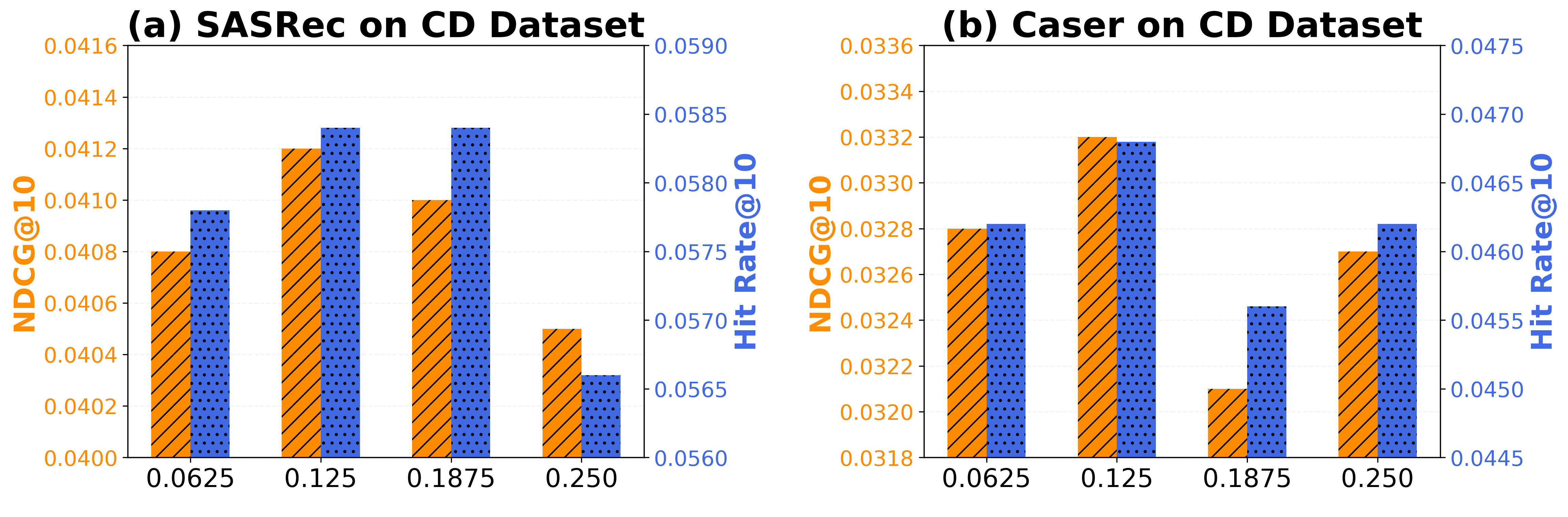}
    \vspace{-7mm}
    \caption{Sensitivity Analysis on Channel Selection Rate}
    \label{fig:sensitivity1}
    \vspace{-5mm}
\end{figure}
Our empirical investigation reveals the impact of threshold parameter $\beta$, which governs the proportion of channels classified into higher precision tiers.
The range of $\beta$ is $\{0.0625$, $0.125$, $0.1875$, $0.25\}$. The larger $\beta$ is, the more channels are seen as sensitive, the larger average bits will be. As illustrated in Figure~\ref{fig:sensitivity1}, performance initially improves, with both SASRec and Caser models demonstrating peak NDCG@10 and Hit Rate@10 at $\beta=0.125$, with an average 3-bit weight. This suggests that the most sensitive channels are effectively captured while maintaining quantization efficiency. When $\beta$ exceeds this value, performance drops noticeably at $\beta=0.1875$ on CD dataset, indicating highly-sensitive channels need to be specially treated for personalization, instead of equally treated as regular channels. Interestingly, a slight recovery occurs at $\beta=0.25$ on CD dataset, attributable to the overall increase in quantization bits. These findings confirm that properly setting the channel selection rate can make a balance between inference efficiency and recommendation performance.


\subsubsection{Sensitivity analysis on bit-width combinations}
\begin{table}[h]
  \caption{Sensitivity Analysis on Bit-width Combinations}
  \vspace{-2mm}
  \label{tab:sensitivity2}
  \resizebox{\linewidth}{!}{
      \begin{tabular}{c|c|cc|cc}
    \toprule[1.5pt]
    \multirow{2}{*}{\textbf{Model}} & \multirow{2}{*}{\textbf{Bit Config}} & \multicolumn{2}{c|}{\textbf{ML-100K}} & \multicolumn{2}{c}{\textbf{Yelp}}  \\
    \cmidrule(lr){3-4}\cmidrule(lr){5-6}
    & & \textbf{NDCG@10} & \textbf{HR@10} & \textbf{NDCG@10} & \textbf{HR@10} \\
    \midrule
    \multirow{2}{*}{\textbf{Caser}} & 2-4-6-8 & 0.0639 & 0.1220 & 0.0143 & 0.0290 \\
    & 2-5-6-7 & 0.0569 & 0.1166 & 0.0140 & 0.0283 \\
    \midrule
    \multirow{2}{*}{\textbf{SASRec}} & 2-4-6-8 & 0.0707 & 0.1463 & 0.0160 & 0.0318 \\
    & 2-5-6-7 & 0.0672 & 0.1368 & 0.0159 & 0.0318 \\
    \bottomrule[1.5pt]
  \end{tabular}
  
  }
\vspace{-3mm}
\end{table}
To investigate the impact of different bit-width combinations, we evaluated two bit-width configurations 
with the same average bit-widths. As shown in Table~\ref{tab:sensitivity2}, the configuration with wider bit-width difference (2-4-6-8) consistently outperforms the configuration with bit-width ranges (2-5-6-7) across both models and datasets. On ML-100K, SASRec with the 2-4-6-8 configuration achieves a 5.21\% improvement in NDCG@10 and 6.94\% in HR@10 compared to the 2-5-6-7 configuration. Similar patterns emerge on the Yelp dataset, though with smaller margins. These results empirically validate that identifying the sensitive channels and assigning higher channel bits will help to preserve the valuable information while improving inference efficiency. When bit-width differentiation is more pronounced, the model better preserves critical information in different sensitivity, demonstrating the effectiveness of our mixed-precision quantization according to parameter sensitivity.

\subsubsection{Evaluation of weight-activation quantization}
\begin{table}[h]
  \vspace{-3mm}
  \caption{Weight-activation Quantization Test}
  \vspace{-2mm}
  \label{tab:weight-activation}
  \resizebox{\linewidth}{!}{
      \begin{tabular}{c|c|c|cc|cc}
    \toprule[1.5pt]
    \multirow{2}{*}{\textbf{Model}} & \multirow{2}{*}{\textbf{w}} & \multirow{2}{*}{\textbf{a}} & \multicolumn{2}{c|}{\textbf{ML-100K}} & \multicolumn{2}{c}{\textbf{Yelp}}  \\
    \cmidrule(lr){4-5}\cmidrule(lr){6-7}
    & & & \textbf{NDCG@10} & \textbf{HR@10} & \textbf{NDCG@10} & \textbf{HR@10} \\
    \midrule
    \textbf{base} & 32 & 32 & 0.0216 & 0.0356 & 0.0132 & 0.0266 \\
    \midrule
    \textbf{ours} & 3 & 32 & 0.0332 &0.0468 & 0.0143 & 0.0290 \\
     \midrule
    \textbf{ours} & 3 & 3 & 0.0327 & 0.0462 & \textbf{0.0146} & \textbf{0.0297} \\
    \bottomrule[1.5pt]
  \end{tabular}
  
  }
\vspace{-2mm}
\end{table}
We further examine the compatibility of our channel-wise mixed-precision weight quantization with activation quantization on backbone Caser, as shown in Table~\ref{tab:weight-activation}. We can observe that our methods outperforms full-precision model with or without activation quantization. When comparing full-precision activations (w=3, a=32) with quantized activations (w=3, a=3), we observe minimal performance degradation in ML-100K, showing only a 1.51\% decrease in NDCG@10 and 1.28\% in HR@10. However, on the Yelp dataset, quantizing both weights and activations actually improves performance by 2.10\% in NDCG@10 and 2.41\% in HR@10. These results demonstrate that our channel-sensitive weight quantization remains effective when combined with activation quantization, offering additional compression benefits with negligible or even positive impact.
\subsubsection{Evaluation of dynamic resource-adaptive deployment}
\begin{table}[h]
  \vspace{-3mm}
  \caption{Adaptation Abilities Test}
    \vspace{-2mm}
  \label{tab:adaptation}
  \resizebox{\linewidth}{!}{
      \begin{tabular}{c|c|cc|cc}
    \toprule[1.5pt]
    \multirow{2}{*}{\textbf{Deployment}} & \multirow{2}{*}{\textbf{Training}} & \multicolumn{2}{c|}{\textbf{CD}} & \multicolumn{2}{c}{\textbf{Yelp}}  \\
    \cmidrule(lr){3-4}\cmidrule(lr){5-6}
    & & \textbf{NDCG@10} & \textbf{HR@10} & \textbf{NDCG@10} & \textbf{HR@10} \\
    \midrule
    \multirow{1}{*}{\textbf{3-bit}} & 3-bit & 0.0332 & 0.0468 & 0.0143 & 0.0290 \\
    \midrule
    \multirow{2}{*}{\textbf{2.5-bit}} & 2.5-bit & 0.0329 & 0.0468 & 0.0138 & 0.0276 \\
    & 3-bit & 0.0332 & 0.0467 & 0.0143 & 0.0290 \\
    \midrule
    \multirow{2}{*}{\textbf{2-bit}} & 2-bit & 0.0329 & 0.0464 & 0.0147 & 0.0295 \\
    & 3-bit & 0.0326 & 0.0459 & 0.0140 & 0.0284 \\
    \bottomrule[1.5pt]
  \end{tabular}
  
  }
\vspace{-2mm}
\end{table}

A key feature of our approach is the ability to dynamically adapt based on available resources. 
Table~\ref{tab:adaptation} demonstrates this capability through two critical aspects. 
First, when deploying a model trained at 3-bit precision to lower bit-widths, we observe the stability in performance. Adapting from 3-bit to 2.5-bit results in no decrease in NDCG@10. When further reducing to 2-bit deployment, the performance degradation remains minimal, with only a 1.81\% decrease in NDCG@10 on CD dataset, achieving elegant performance degradation. Second, compared to dedicated training at target precision, our approach achieves comparable performance. For instance, a model trained directly at 2.5-bit precision achieves 0.0329 NDCG@10 on CD, while our adapted 3-bit model achieves 0.0332. These results confirm CHORD's superior performance under varying resource constraints, enabling devices to adapt with real-time resource availability, which is particularly valuable for real-world deployment.

\subsection{Visualization}
\begin{figure}
    \vspace{-2mm}
    \centering
    \includegraphics[width=1\linewidth]{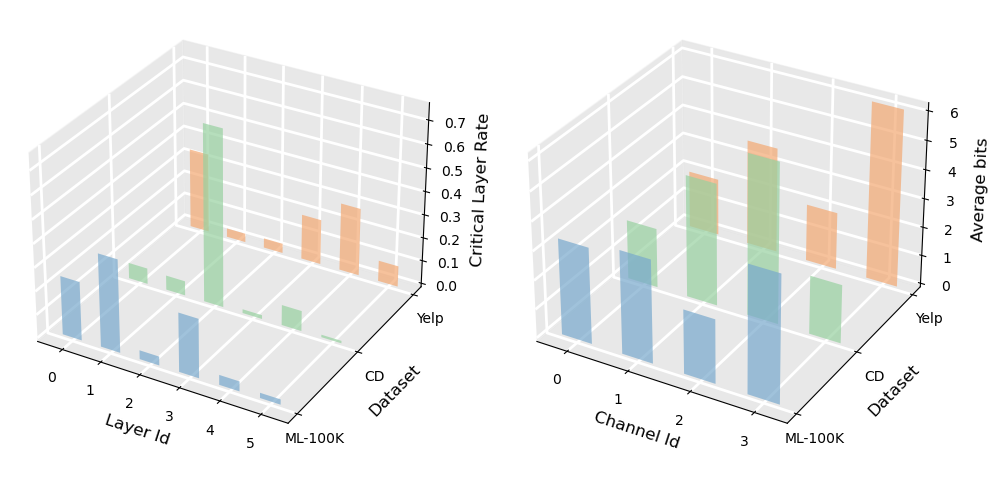}
    \vspace{-8mm}
    \caption{Visualization of the personalized quantization strategy: The left subplot demonstrates the distribution of layers identified as most critical. The right subplot displays the average bit allocation per channel for users in the 0th layer.}
    \vspace{-4mm}
    \label{fig:visualization}
    \vspace{-2mm}
\end{figure}
Figure \ref{fig:visualization} presents the personalized mixed-precision quantization models in visualization. The left subplot demonstrates the distribution of layers identified as most critical. We observe that the most sensitive layers vary significantly across both users and datasets. For Yelp and ML-100K datasets, half of the layers have similar potential to be most sensitive, further confirming a unified mixed-precision strategy is not good enough to capture the optimal model for users. The right subplot shows the average bit allocation per channel in layer 0th. Similar to the left figure, we observe distinct allocation patterns for each user-dataset combination. This visualization validates that our channel-wise mixed-precision quantization successfully identifies user-specific features and tailors bit allocation accordingly, providing personalized and efficient quantization. 

\section{Conclusion}
In this work, we introduce an efficient framework named CHORD, leveraging on-device mixed-precision quantization to simultaneously achieve personalization and resource-adaptive deployment. To identify channels critical for maintaining recommendation performance, we develop multi-level sensitivity extractors on the cloud, while designing a user profiling generator on the device. CHORD generates channel-wise quantization strategy based on user behaviors, considering layer, filter, and element level importance. Additionally, we encode the customized strategy into 2 bits per channel, enhancing communication efficiency. 
Extensive experiments demonstrate the  accuracy, efficiency, and adaptability of CHORD, highlighting the framework’s potential for practical applications. Future work will focus on integrating large language models to refine the collaboration mechanisms and improve personalized recommendation. 

\begin{acks}
This project is supported by the National Science and Technology Major Project (2022ZD0119100), the National Natural Science Foundation of China (No. 62402429, U24A20326, 62441236), the Key Research and Development Program of Zhejiang Province (No. 2025C01026, 2024C03270), the Ningbo Yongjiang Talent Introduction Programme (2023A-397-G), and the Young Elite Scientists Sponsorship Program by CAST (2024QNRC001). The author gratefully acknowledges the support of the Zhejiang University Education Foundation Qizhen Scholar Foundation.
\end{acks}

\bibliographystyle{ACM-Reference-Format}
\bibliography{sample-base}

\end{document}